\documentclass[journal]{IEEEtran}

\usepackage{ifpdf}

\usepackage{cite}

\usepackage{mytexpckgs}

\newlength\mylen

\newcolumntype{N}{>{\centering\arraybackslash}m{3cm}}
\newcolumntype{W}{>{\centering\arraybackslash}m{7cm}}

\DeclareMathOperator*{\argmin}{arg\,min}

\begin{document}
%
\title{Multi-Step Prediction of Dynamic Systems with Recurrent Neural Networks}

\author{Nima~Mohajerin,~\IEEEmembership{Member,~IEEE,}
        Steven~L.~Waslander,~\IEEEmembership{Member,~IEEE,}}

%
%

\markboth{Multi-Step Prediction of Dynamic Systems with RNNs, N. Mohajerin, S. L. Waslander}%
{Shell \MakeLowercase{\textit{et al.}}: Bare Demo of IEEEtran.cls for IEEE Journals}

\maketitle

\begin{abstract}
Recurrent Neural Networks (RNNs) can encode rich dynamics which makes them suitable for modeling dynamic systems. To train an RNN for multi-step prediction of dynamic systems, it is crucial to efficiently address the state initialization problem, which seeks proper values for the RNN initial states at the beginning of each prediction interval. In this work, the state initialization problem is addressed using Neural Networks (NNs) to effectively train a variety of RNNs for modeling two aerial vehicles, a helicopter and a quadrotor, from experimental data. It is shown that the RNN initialized by the NN-based initialization method outperforms the state of the art. Further, a comprehensive study of RNNs trained for multi-step prediction of the two aerial vehicles is presented. The multi-step prediction of the quadrotor is enhanced using a hybrid model which combines a simplified physics-based motion model of the vehicle with RNNs. While the maximum translational and rotational velocities in the quadrotor dataset are about 4 m/s and 3.8 rad/s, respectively, the hybrid model produces predictions, over 1.9 second, which remain within 9 cm/s and 0.12 rad/s of the measured translational and rotational velocities, with 99\% confidence on the test dataset.
\end{abstract}
\begin{IEEEkeywords}
Recurrent Neural Networks, Nonlinear Dynamic System Modeling, State Initialization, Multi-step Prediction
\end{IEEEkeywords}

%
\IEEEpeerreviewmaketitle

\section{Introduction}
\IEEEPARstart{M}{ulti-step} prediction of dynamic motion remains a challenging problem.  The simplifying assumptions required in the process of modeling dynamic systems lead to \emph{unmodeled dynamics}, which in turn lead to repeatable and systematic prediction errors. When the prediction is required over very short periods into the future, the unmodeled dynamics may cause negligible error, however, using the model over longer prediction horizons, the unmodeled dynamics can lead to drastic growth in the prediction error over time. In the discrete-time domain, a prediction required for one time step into the future is referred to as a single-step prediction, and a prediction many steps into the future is referred to as a multi-step prediction.

In this paper, the multi-step prediction of mobile robotic systems is considered. Recurrent Neural Networks (RNNs) are mainly employed to learn the dynamics of two aerial vehicles, a helicopter and a quadrotor, from experimental data. The helicopter dataset belongs to the Stanford Helicopter platform\footnote{http://heli.stanford.edu/dataset/} which has been used in Apprenticeship Learning~\cite{Abbeel2010} and single-step prediction and modeling of the platform~\cite{Abbeel2015}. The quadrotor dataset, however, has been specifically collected for this work. 

Multi-step prediction has many applications in state estimation, simulation and control~\cite{Parlos2000, Bone2002}. For example, model based control schemes, such as Model Predictive Control (MPC)~\cite{Maciejowski2002}, can extensively benefit from an accurate long-term prediction. Accurate multi-step predictions allow for a slower update rate of the MPC, reducing the overall computational burden while maintaining smoothness and accuracy of the resulting control system response. As another example, in a moving vehicle when some measurements, such as GPS readings, are temporarily unavailable, a multi-step prediction can account for the missing measurements and approximate the system position and speed over the blackout period.

The classic method of modeling, i.e., modeling from first principles (or white-box methods), suffer from two major difficulties. First, the developed model will contain many parameters which describe the system physical characteristics (mass, drag coefficient, etc.) and must be properly identified prior to using the model. Second, many properties of the system might be too difficult to model explicitly, such as the vortex-ring effect on a quadrotor vehicle~\cite{Hoffmann2011}. Identifying the parameters of a model can be expensive. For instance, measuring the blade drag coefficient of a quadrotor needs a wind tunnel. Moreover, by changing the system physical properties slightly, the model should be adapted accordingly, which may involve new measurements and cumbersome tests.

On the other hand, learning-based (or black-box) modeling relies on the observations from the system to decipher complex nonlinearities acting on the system. There are several black-box architectures, such as polynomial models (the Wiener model, Voltera series, etc.), Fuzzy models~\cite{Yager2012, Oviedo2006} and Neural Networks~\cite{Haykin_Book_1999, Nelles2013}. Regardless of the method, a black-box model has many degrees-of-freedom (DOF), depicted as parameters or weights, that should be found based on a set of input-output observations from the system. The search to find the appropriate values for the model parameters is usually done through an optimization process and, since many black-box models are Machine Learning (ML) methods, the parameter optimization process is frequently referred to as a \emph{learning} process. 

In recent years, ML has been going through rapid development. Deep Learning (DL)~\cite{Hinton2006:1,Hinton2006:2} is revolutionizing the ML field and solving problems that could not have been solved before, such as speech recognition~\cite{Hinton2012} and object detection and classification in images~\cite{Badrinarayanan2015,Szegedy2017}. Three main components propel this revolution: i) the newly available hardware capable of performing efficient parallel processing on large numerical models, ii) methods to train these models and iii) availability of large-scale datasets. Because of the third reason, the applications of deep learning have been mainly focused on natural language processing and image classification for which such datasets are readily available. However, DL methods can be also used in modeling and control of robotic systems. In fact, in mobile robotics, where the robot is deployed in an unstructured environment with noisy and uncertain sensory information, learning from observation can deal with problems that are either too difficult or too expensive to handle with classical methods. This work demonstrates some of the capabilities and applicability of ML methods in such applications. Particularly, in this work we show how to employ RNNs to accurately predict the behavior of the two robotic systems using input information only, many steps into the future.

Using RNNs in multi-step prediction requires a proper state initialization, that is, assigning proper initial values to the neuron outputs at the first step of prediction. The underlying reason is that the RNN recursively updates itself throughout the prediction horizon and therefore, exhibits dynamics. Like any other dynamic system, the (transient) response of an RNN in general depends heavily on its initial condition. For an RNN with hidden layers, the common approach is to initialize the hidden neuron outputs to zero (or random) values and run the network until the effect of the initial values washes out~\cite{Zimmermann2012, Jaeger2002}. Some of the drawbacks in the washout method are:
\begin{itemize}
	\item The washout period is not fixed and hard to determine a priori, during which the network does not produce a reliable prediction.
	\item The network may become unstable during the washout period, leading to prolonged or even failed training sessions.
	\item In the training process, the input sequence used in washout does not contribute to the learning process.
\end{itemize}
The third drawback listed above is more severe when the dataset collection process is expensive, which is often the case with experimental data acquisition in the robotics domain.

In this work, the RNN state initialization problem is carefully studied. An NN-based initialization method, previously proposed in~\cite{Mohajerin2017_1}, is revisited and expanded. It is shown that the network initialized with the NN-based method significantly outperforms the same network initialized by the washout method. After establishing the effectiveness and efficiency of our NN-based initialization method, the results of black-box modeling of the two aerial vehicles are demonstrated and studied. To further improve the quadrotor model, the hybrid model in~\cite{Mohajerin2017_2} is revisited. This work embodies the capability of neural networks in learning unmodeled dynamics of a real and challenging robotic system for multi-step prediction, for the first time, and may serve as a basis for future development and application of more sophisticated learning-based methods in modeling, prediction and control of such systems. 

\section{Background and Literature Review}

Feed-Forward Neural Networks (FFNNs) have been used extensively in modeling and control of dynamic systems~\cite{Akpan11,Nerrand1994,Delgado95,Baruch09,Atuonwu10,AlSeyab08-1}. In a control problem, they may be employed as a modelling part of a controller in a Lyapunov design approach. In this method using a Lyapunov function, the equations for evolution of the neural network weights are extracted so that the closed loop controller stabilizes the system~\cite{Madani2008,Boudjedir2012,Dierks2010, Nicole2011}. Since FFNNs lack exhibiting dynamics, they are mainly used as single-step predictor or compensator.

RNNs possess dynamics and are universal approximators for reconstructing state-space trajectories of dynamic systems~\cite{Funahashi93, Jin1995, Schafer2006}, which make them suitable candidate models for multi-step prediction problem. In~\cite{Funahashi93}, it is shown that any finite-time trajectory of a dynamic system can be approximated by some RNNs to any desired level of accuracy given a \emph{proper initial state}. This result is extended to discrete RNNs in~\cite{Jin1995}. This is another aspect to reinforce the importance of a proper state initialization for RNN in modeling dynamic systems.

Nonlinear Auto-Regressive (NAR) models are classic tools to model dynamic systems~\cite{Nelles2013, Narendra90, Chen1990}. In~\cite{Narendra90}, Narendra et al. devised methods to use Multi-Layer Perceptrons (MLPs) in the Non-linear Autoregressive eXogenous (NARX) framework. In a discrete-time fashion, NARX framework implements a dynamic system whose output at any given time, $y_k$, is a function of the input at that time, $u_k$, and the system output and input at previous time steps,
\begin{equation*}
y_k=f\big{(}y_{k-1},\dots,y_{k-\tau_y}, u_k,u_{k-1},\dots,u_{k-\tau_u}\big{)},
\end{equation*}
where the length of the memory buffers, i.e., $\tau_u$ and $\tau_y$, are usually given or determined through a hyper-parameter optimization process. The function $f(.)$ can be realised by various methods. In~\cite{Narendra90}, the function $f(.)$ is realized by an MLP. To avoid confusion, the method to implement this function is added to the NARX abbreviation as a suffix. For instance, if $f(.)$ is realised by an MLP then the architecture will be referred to as NARX-MLP. A NARX-MLP is essentially an MLP equipped with buffers and feedback connections. Hence, it can be classified as an RNN.

The NARX-MLP architectures are often trained via a \emph{Series-Parallel}~\cite{Narendra90} mode which uses the network target values instead of the network past outputs as the delayed version(s) of the network output. This method is also known as \emph{teacher forcing}~\cite{Narendra90}. Clearly, this mode converts the NARX architecture into a feedforward one which therefore loses the main advantage of an RNN and limits the ability of the method to represent dynamical systems accurately. On the other hand, training NARX-MLP in a closed-loop form (Parallel mode) to model dynamic systems can be difficult due to numerical stability issues in the calculation of the gradient for learning based optimization~\cite{Mohajerin2014_1}.

One alternative to NARX model is to define an internal state, $x_k$, and use a one step memory buffer, 
\begin{align*}
x_k=&f\big{(}x_{k-1},y_{k-1},u_k\big{)},\\
y_k=&g\big{(}x_k\big{)}.
\end{align*}
This architecture is an example of a Recurrent Multilayer Perceptron (RMLP). An RMLP is made by one (or a few) locally recurrent layers of sigmoid neurons~\cite{KolenKremer}. RMLPs have been used in a number of dynamic system identification and modelling problems, such as a heat exchanger~\cite{Parlos94}, engine idle operation~\cite{Li2002} and wind power prediction~\cite{Shuhui03}. 

It is not clear whether using RMLPs is more advantageous than NARX-MLPs. However, in~\cite{Kumar2006} it is shown that NARX-MLPs, in a single-step prediction scenario, outperform RMLPs in modelling a real helicopter. NARX RNNs have been extensively studied~\cite{Siegelmann97,Tsungnan96} and used in various modelling and identification problems~\cite{Basso05,Anderson10,Zhan14}. In~\cite{Wu2014}, a Radial Basis Function (RBF) network in a NARX architecture, i.e., NARX-RBF, is used to model and control a quadrotor with three horizontal and one vertical propeller. In~\cite{Taha2010}, another form of NARX-RBF is employed and trained using Levenberg-Marquardt (LM) optimization algorithm to model a small-scale helicopter. Both approaches employ teacher forcing. 

Recently, Abbeel et al. used Rectified Linear Units neural networks to model dynamics of a real helicopter~\cite{Abbeel2015}. Although they have not used RNNs, their dataset is also used in this work to assess the performance of RNNs. In~\cite{Zhang2016}, a deep neural network, trained by a Model-Predictive Control (MPC) policy search, is used as a policy learner to control a quadrotor. The network generates one step policies and has a feed-forward architecture. In~\cite{Ogunmolu2016}, a few NN architectures, such as MLPs, RMLPs, Long-Short-Term-Memory (LSTM) and Gated Recurrent Unit cells are compared against each other in single step prediction of a few small robotic datasets. In~\cite{Lenz2015}, a hybrid of recurrent and feed-forward architectures is used to learn the latent features for MPC of a robotic end-effector to cut 20 types of fruits and vegetables. Although the authors use recurrent structure, they also state that using their \emph{Transforming Recurrent Units} (TRUs) in a multi-step prediction scheme results in ``instability in the predicted values", so they use their proposed network as a one-step predictor. However, the recurrent latent state helps to improve the predictions.

\section{Problem Formulation}
\label{sec:problem_formulation}
In this section, the multi-step prediction problem is defined for a general non-linear dynamic system. As discussed earlier, regardless of the system to be modeled by RNNs, a proper state initialization is required and therefore, the state initialization problem is also defined. Note that for practical purposes we are working in the discrete time domain.

\subsection{Multi-Step Prediction with Recurrent Neural Networks}
Consider a dynamic system $\mathcal{S}_n^m$ with $m$ input and $n$ output dimensions. The system input and output at a time instance, $k$, is denoted by  $\mathbf{u}_k\in\mathbb{R}^m$ and $\mathbf{y}_k\in\mathbb{R}^n$, respectively. We assume that both the input and output are measurable at all $k$s. Consider an input sequence of length $T$ starting at a time instance $k_0+1$, $\mathbf{U}_{k_0+1,T}\in\mathbb{R}^m\times\mathbb{R}^T$,
\begin{equation}
	\mathbf{U}_{k_0+1,T} =
		\begin{bmatrix}
			\mathbf{u}_{k_0+1} & \mathbf{u}_{k_0+2} &\hdots & \mathbf{u}_{k_0+T}
		\end{bmatrix}.
	\label{eq:input_seq}
\end{equation}
The system response to this input is an output sequence denoted by $\mathbf{Y}_{k_0+1,T}\in\mathbb{R}^n\times\mathbb{R}^T$,
\begin{equation}
	\mathbf{Y}_{k_0+1,T} =
		\begin{bmatrix}
			\mathbf{y}_{k_0+1} & \mathbf{y}_{k_0+2} &\hdots & \mathbf{y}_{k_0+T}
		\end{bmatrix}.
	\label{eq:output_seq}
\end{equation}
The multi-step prediction problem seeks an accurate estimate of the system output over the same time-horizon, $\mathbf{\tilde{Y}}_{k_0+1,T}\in\mathbb{R}^n\times\mathbb{R}^T$,
\begin{equation}
	\mathbf{\tilde{Y}}_{k_0+1,T} =
		\begin{bmatrix}
			\mathbf{\tilde{y}}_{k_0+1} & \mathbf{\tilde{y}}_{k_0+2} &\hdots & \mathbf{\tilde{y}}_{k_0+T}
		\end{bmatrix},
	\label{eq:output_seq2}
\end{equation}
which minimizes a Sum-of-Squares Error measure (SSE) over the prediction length, $T$,
\begin{align}
	\label{eq:rnn_cost}
	&L =\sum_{k=k_0+1}^{k_0+T}\mathbf{e}_k^\top\mathbf{e}_k\\
	&\mathbf{e}_k = \mathbf{y}_k - \mathbf{\tilde{y}}_k.
\end{align} 

An RNN with $m$ input and $n$ output, $\mathcal{R}^m_n$, is a dynamic system. At each time instance $k$, feeding the input element $\mathbf{u}_k$ to the RNN, it evolves through two major steps: 1) state update and 2) output generation,
\begin{subequations}
	\begin{align}
		\mathbf{x}_k(\boldsymbol{\theta}) =& \mathbf{f}\big{(}\mathbf{x}_{k-1}(\boldsymbol{\theta}), \mathbf{u}_k\big{)},\\
		\mathbf{\tilde{y}}_k(\boldsymbol{\theta}) =& \mathbf{g}\big{(}\mathbf{x}_k(\boldsymbol{\theta}), \mathbf{u}_k\big{)},
	\end{align}
	\label{eq:rnn_evolution_general}
\end{subequations}
where $\mathbf{\tilde{y}}_k(\boldsymbol{\theta})$ and $\mathbf{x}_k(\boldsymbol{\theta})$ are the RNN output vector and state vector, respectively, at time $k$. The vector $\boldsymbol{\theta}\in\mathbb{R}^q$ encompasses the network weights and $q$ depends on the architecture of the RNN. The function $\mathbf{f}(.)$ is defined either explicitly (RMLPs)~\cite{KolenKremer} or implicitly (LSTMs)~\cite{Schmidhuber97} as are discussed in Section~\ref{sec:frameworks}. For $\mathbf{g}(.)$, however, we choose a linear map because modeling a dynamic system by RNN is a regression problem. Note that the network states and/or output may also depend on a \emph{history} of the right-hand-side values in the equations~\eqref{eq:rnn_evolution_general}.\\

In the discrete domain, feedback connections require some form of a memory buffer. In an RNN, \emph{states} are the buffered values since they determine the network output knowing the input and network functions. Depending on the feedback source, there are two types of states: 1) output-states, $\mathbf{o}_k$, which encompasses feedbacks from the network output, and 2) internal-states, $\mathbf{s}_k$,  which encompasses feedbacks from within the network. Therefore the RNN state vector is,
\begin{equation}
	\mathbf{x}_k = 
		\begin{bmatrix}
			\mathbf{o}_k \\
			\mathbf{s}_k
		\end{bmatrix}\in\mathbb{R}^s.
	\label{eq:states}
\end{equation}
where $s$ is the states count. Therefore, we can rewrite equations~\eqref{eq:rnn_evolution_general} using single-step buffers,
\begin{subequations}
	\begin{align}
		\mathbf{o}_k(\boldsymbol{\theta}) = &\mathbf{\tilde{y}}_{k-1}(\boldsymbol{\theta}),\\
		\mathbf{s}_k(\boldsymbol{\theta}) =& \mathbf{f}(\mathbf{x}_{k-1}(\boldsymbol{\theta}), \mathbf{u}_k),\\
		\mathbf{\tilde{y}}_k(\boldsymbol{\theta}) =& \mathbf{g}(\mathbf{x}_k(\boldsymbol{\theta}), \mathbf{u}_k).
	\end{align}
	\label{eq:rnn_evolution}
\end{subequations}

Using RNNs to address multi-step prediction problem, we seek an RNN which given any input sequence $\mathbf{U}_{k_0+1,T}$, produces an output sequence $\mathbf{\tilde{Y}}_{k_0+1,T}(\boldsymbol{\theta})$ which minimizes the Mean SSE (MSSE) cost over the prediction interval $[k_0+1,k_0+T]$,
\begin{align}
	\label{eq:rnn_ideal_cost}
	&L(\boldsymbol{\theta}) = \frac{1}{T}\sum_{k=k_0+1}^{k_0+T}\mathbf{e}_k(\boldsymbol{\theta})^\top\mathbf{e}_k(\boldsymbol{\theta})\\
	&\mathbf{e}_k(\boldsymbol{\theta}) = \mathbf{y}_k - \mathbf{\tilde{y}}_k(\boldsymbol{\theta}).
\end{align} 
where $\mathbf{y}_k$ is the system output at time $k\in[k_0+1,k_0+T]$ to the input $\mathbf{u}_k\in\mathbf{U}_{k_0+1,T}$.
Therefore, the solution to the multi-step prediction problem is an RNN which minimizes $L$ for all possible input-output sequences,
\begin{equation}
	\boldsymbol{\theta}^* = \operatornamewithlimits{\argmin}_{\boldsymbol{\theta}}\big{(}L(\boldsymbol{\theta})\big{)}
	\label{eq:MS_solution}
\end{equation}
The optimization in equation~\eqref{eq:MS_solution} is practically impossible because there can be infinite input-output sequences. In practice, a dataset is collected by measuring the system input and output and a numerical optimization is carried out to find \emph{a} minimum of the total cost,
\begin{equation}
	\begin{aligned}
		L_{pred}(\boldsymbol{\theta}) = &\frac{1}{D}\sum_{i=1}^{D}L_i(\boldsymbol{\theta}) \\
		=& \frac{1}{TD}\sum_{i=1}^{D}\sum_{k=k_0+1}^{k_0+T}\mathbf{e}_{i,k}(\boldsymbol{\theta})^\top\mathbf{e}_{i,k}(\boldsymbol{\theta}),
	\end{aligned}
	\label{eq:L_pred}
\end{equation}
where $D$ is the dataset size. The datasets employed in the optimization process are comprised of time-series samples in the form of input-output tuples,
\begin{equation}
	\mathbb{D}=\Big{\{}s_i=\big{(}\mathbf{U}_i(T),\mathbf{Y}_i(T)\big{)}\Big{\}}.
\end{equation}
where $T$ indicates that all samples are of the same length.

There are many tricks and tweaks to enhance RNN training which some of them are used in this paper and will be mentioned in Section~\ref{sec:results}. For detailed discussions refer to~\cite{Jaeger2002} and~\cite{Zimmermann2012}.

\subsection{State Initialization in Recurrent Neural Networks}
Given an initial state, $\mathbf{x}_{k_0}$, we can rewrite the RNN input-output equation as a sequence-to-sequence map,
\begin{equation}
	\mathbf{\tilde{Y}}_{k_0+1,T} = \mathbf{F}\big{(}\mathbf{x}_{k_0},\mathbf{U}_{k_0+1,T}\big{)}.
	\label{eq:seq2seq}
\end{equation}
The function $\mathbf{F}:\mathbb{R}^s\times\mathbb{R}^m\times\mathbb{R}^T\to\mathbb{R}^n\times\mathbb{R}^T$, symbolizes the operations taking place sequentially inside the RNN, defined by~\eqref{eq:rnn_evolution}.

From~\eqref{eq:seq2seq}, it is evident that the initial states play a key role in the immediate response of the RNN. Therefore, to have an accurate estimate one should properly \emph{initialize} the RNN, i.e., set the initial states of the RNN, $\mathbf{x}_k$, to values that \eqref{eq:L_pred} is minimized. Note that minimizing \eqref{eq:L_pred} also requires training the RNN. Therefore, RNN state initialization should be considered as a part of RNN training which should be repeatable in the testing (generalization tests) as well. We will explain this insight in more details in Section~\ref{sec:state_init_methods}.

\section{Solutions to RNN state initialization problem}
\label{sec:state_init_methods}
In the context of dynamic system modeling with RNNs, the function that produces the network output, i.e., $\mathbf{g}(.)$ in~\eqref{eq:rnn_evolution}, is a linear mapping. For a prediction generated at the time-instance $k$, the output is,
\begin{equation}
	\mathbf{\tilde{y}}_k = \mathbf{A}\mathbf{x}_k + \mathbf{B}\mathbf{u}_k,
	\label{eq:output}
\end{equation}
where $\mathbf{A}\in\mathbb{R}^n\times\mathbb{R}^s$ and $\mathbf{B}\in\mathbb{R}^n\times\mathbb{R}^m$ are the output layer weights and their elements are included in the weight vector $\boldsymbol{\theta}$, hence, we have dropped $\boldsymbol{\theta}$. Using \eqref{eq:states} to expand \eqref{eq:output} and letting $k=k_0$ we have,
\begin{subequations}
	\begin{align}
		\mathbf{\tilde{y}}_{k_0}&= \mathbf{A}_s\mathbf{s}_{k_0} + \mathbf{A}_o\mathbf{o}_{k_0} + \mathbf{B}\mathbf{u}_{k_0},\\
	 	\mathbf{A}&=\big{[}\mathbf{A}_s \hspace{5pt} \mathbf{A}_o\big{]}.
	\end{align}
	\label{eq:output_div}
\end{subequations}

According to \eqref{eq:rnn_evolution}, at each time-instance $k\in[k_0+1, k_0+T]$ the states, $\mathbf{x}_k$, must be updated prior to generating the output, $\mathbf{\tilde{y}}_k$, which requires the knowledge of $\mathbf{x}(k-1)$. Based on the universal approximation property, there exists an RNN\textsuperscript{*} which can approximate the system to be modelled with $\boldsymbol{\epsilon}_k$ error accuracy, where $\boldsymbol{\epsilon}_k$ is the prediction error at time $k$ and can be decreased infinitesimally for all $k$~\cite{Jin1995}. Therefore, Equation~\eqref{eq:output_div} for the RNN\textsuperscript{*} becomes,
\begin{equation}
	\mathbf{\tilde{y}}^*_{k_0}= \mathbf{A}^*_s\mathbf{s}^*_{k_0} + \mathbf{A}^*_o\mathbf{o}^*_{k_0} + \mathbf{B}^*\mathbf{u}_{k_0},
\end{equation}
and using the prediction error $\boldsymbol{\epsilon}_{k_0}$, 
\begin{equation}
	\begin{aligned}
		\mathbf{y}_{k_0}+\boldsymbol{\epsilon}_{k_0}= \mathbf{A}^*_s\mathbf{s}^*_{k_0} +\mathbf{A}^*_o\mathbf{o}_{k_0-1}+\mathbf{A}^*_o\boldsymbol{\epsilon}_{k_0-1} +\mathbf{B}^*\mathbf{u}_{k_0}.
		\end{aligned}
	\label{eq:output_div*}
\end{equation}
Letting $\boldsymbol{\epsilon}_k\to\mathbf{0}$ for all $k$,
\begin{equation}
	\mathbf{A}_s^*\mathbf{s}^*_{k_0-1} \approx \mathbf{c}^*,
\end{equation}
where,
\begin{equation*}
	\mathbf{c}^* = \mathbf{y}_{k_0} - \mathbf{A}_o^*\mathbf{y}_{k_0-1} - \mathbf{B}^*\mathbf{u}_{k_0}.
\end{equation*}
Note that the weights are known at the time of state initialization. However, the RNN\textsuperscript{*} is not necessarily known. Therefore, during training an RNN, the state initialization problem for system identification boils down to minimizing the following cost,
\begin{equation}
	L_{si}=|\mathbf{A}_s\mathbf{s}_{k_0} - \mathbf{c}|,
	\label{eq:state_init_cost}
\end{equation}
with respect to $\mathbf{s}_{k_0}$ subject to $a\leq\mathbf{s}_{k_0}\leq b$. The constraints should enforce the initial states to remain within the range of the function which generates the hidden-states. For example if the states are generated by a $\tanh(.)$ then $a=-1, b=+1$.
Ideally we want the $L_{si}=0$, however, since the initialization has to be carried out in both during training and testing phases, the exact solution may lead to overfitting as will be described later in this section.

In this section , three methods of state initialization in RNNs are studied. The first method, which is also referred to by the term \emph{washout} is described in~\cite{Jaeger2002} and perhaps is the most commonly used. The second method is based on optimizing the initial states along with the network weights~\cite{Becerra2002}. The third method employs NNs and has been proposed previously by the authors~\cite{Mohajerin2017_2} and is expanded here.

\subsection{RNN State Initialization: Washout}
The washout method is based on the idea that running an RNN for a period of time steps attenuates (washes out) the effect of the initial state values. Therefore, one can set the initial states to zero~\cite{Jaeger2002}, or a random value~\cite{Zimmermann2012}, run the RNN for a length of time until the effect of the initial values wears off.

Since there is no deterministic approach to obtain the washout period, it has to be treated as a hyper parameter and reduces the speed of the learning process. Additionally, since RNNs are dynamic systems, during the training process, they may temporarily experience instability. While over the entire learning process the stability of RNNs is chained to the stability of the system being learned, running an unstable instance of an RNN during training may result in extremely large cost values which cause the learning curve to diverge. To avoid such "blow-ups", extra measures seem necessary. In this paper we employ washout as it is stated in~\cite{Jaeger2002}.

\subsection{RNN State Initialization: Optimization}
Another proposed approach to initialize an RNN is to augment the initial states $\mathbf{s}_k$ to the weight vector $\boldsymbol{\theta}$ and carry out the learning process with respect to the augmented weight vector~\cite{Becerra2002}. Although this approach may address the state initialization problem during the training phase, it does not provide a mechanism to generate the initial state values in general, after the training is finished. For special cases (for instance in~\cite{Becerra2002}) it may be possible to obtain the initial hidden state values by running an architecture or problem specific optimization process. Since a general approach that is applicable to any RNN regardless of the architecture is more appealing, this method is not desirable for our purposes.

\subsection{RNN State Initialization: NN-based}
Consider the ideal RNN, RNN\textsuperscript{*}, where the network output differs from the desired output infinitesimally. We can write, 
\begin{subequations}
	\begin{align}
		\mathbf{o}_k = &\mathbf{y}^*_{k-1} \approx \mathbf{y}_{k-1}, \\
		\mathbf{s}_k = & \mathbf{f}(\mathbf{x}_{k-1}, \mathbf{u}_k).
	\end{align}
	\label{eq:RNN_states_dynamics}
\end{subequations}
Equations~\eqref{eq:RNN_states_dynamics} govern the dynamics of the RNN\textsuperscript{*} \emph{states}. To approximate this mapping, it is possible to employ NNs. In~\cite{Mohajerin2017_2}, we have proposed an auxiliary FFNN to produce the RNN~\emph{initial state values}, receiving a short history of the system input and output. To avoid confusion, the auxiliary network will be referred to as the \emph{initializer} and the RNN which performs the prediction as the \emph{predictor}. 

\begin{figure}[htp!]
	\centering
		\includegraphics[scale=0.46]{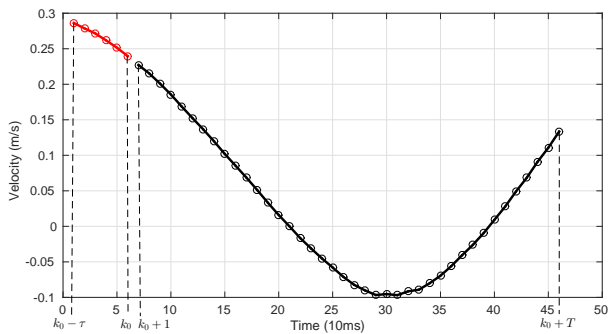}
	\caption[Dividing a data sample into initialization and prediction segments.]{Dividing a data sample into initialization (red) and prediction (black) segments. Each small circle is one measurement from the continuous signal. In this figure, $\tau = 6$ and $T=40$. }
	\label{fig:data_sample}
\end{figure}

The idea is to divide the data samples into two segments; the first segment is used as the input to the initializer, which initializes the predictor states, and the second one is used to train the whole network, i.e., the initializer-predictor pair. The number of steps in the prediction and initialization segment will be referred to as the prediction and initialization length and denoted by $T$ and $\tau$, respectively, as illustrated in Figure~\ref{fig:data_sample}. The total length of the training sample is therefore $T_{tot}=\tau+T$.

The desired values for the output of the initializer network, i.e., the initial RNN state values, are unknown. However,~\eqref{eq:state_init_cost} proposes a penalty on the initializer network output. Therefore, the initializer-predictor pair will be trained on the following cost,
\begin{equation}
		L_{tot} = \alpha L_{pred} + \beta L_{si} 
	\label{eq:MS_Pred_Total_Cost}
\end{equation}
where the prediction error, $L_{pred}$ and $L_{si}$ are defined in~\eqref{eq:L_pred} and \eqref{eq:state_init_cost}, respectively, and the coefficients $\alpha$ and $\beta$ can be used to balance between the two costs. Without loss of generality, $\alpha$ and $\beta$ are set to one in this work. Nevertheless, they can be treated as hyper-parameters and tuned to achieve desired performance, if necessary.

\textbf{MLP Initializer Network}:
An MLP can be employed as the initializer network, which receives a history of the measurements from the system and produces the predictor initial states. 
\begin{equation}
	\begin{aligned}
		\mathbf{x}_{k_0} = \boldsymbol{\zeta}\big{(}&\mathbf{u}_{k_0-\tau},\mathbf{u}_{k_0-\tau+1},...,\mathbf{u}_{k_0},& \\
		&  \mathbf{y}_{k_0-\tau},\mathbf{y}_{k_0-\tau+1},..., \mathbf{y}_{k_0}\big{)}.
	\end{aligned}
	\label{eq:init_state_mapping}
\end{equation}

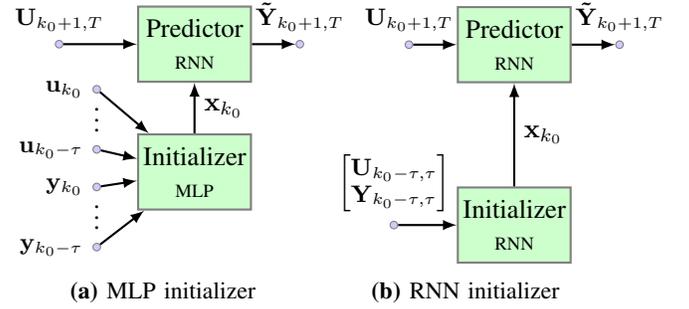
\begin{figure}
	\begin{center}
		\begin{subfigure}{.23\textwidth}
			\begin{tikzpicture}
				\node[NN,minimum height=1cm,minimum width=1.5cm,text width=1.5cm, align=center] (PREDICTOR) at (1,0) []{Predictor \\ \scriptsize{RNN}};
				\node[NN,minimum height=1cm,minimum width=1.5cm,text width=1.5cm, align=center] (Initializer) at (1,-1.7) []{Initializer \scriptsize{MLP}}
				edge[post] node[right]{{$\mathbf{x}_{k_0}$}} (PREDICTOR);
				\node[sig] (U)   at (-0.8,0) [label=above:\small{$\mathbf{U}_{k_0+1,T}$}] {};
				\node[sig] (Y)   at (2.4,0)    [label=above:\small{$\mathbf{\tilde{Y}}_{k_0+1,T}$}] {};
				\node[sig] (u_1) at (-0.3,-0.6) [label=left:\small{$\mathbf{u}_{k_0}$}] {};
				\node[]    (dots)at (-0.3,-0.9) {$\vdots$};
				\node[sig] (u_n) at (-0.3,-1.4) [label=left:\small{$\mathbf{u}_{k_0-\tau}$}] {};
				\node[sig] (y_1) at (-0.3,-1.9) [label=left:\small{$\mathbf{y}_{k_0}$}] {};
				\node[]    (dots)at (-0.3,-2.2) {$\vdots$};
				\node[sig] (y_n) at (-0.3,-2.7) [label=left:\small{$\mathbf{y}_{k_0-\tau}$}] {};
				\draw[->,>=latex,thick] (U) -- (PREDICTOR) ;
				\draw[->,>=latex,thick] (PREDICTOR) -- (Y) ;
				\draw[->,>=latex,thick] (u_1) -- (Initializer) ;
				\draw[->,>=latex,thick] (u_n) -- (Initializer) ;
				\draw[->,>=latex,thick] (y_1) -- (Initializer) ;
				\draw[->,>=latex,thick] (y_n) -- (Initializer) ;
			\end{tikzpicture}
			\caption{MLP initializer}
			\label{fig:InitializationArch1}
		\end{subfigure}				
		\begin{subfigure}{.20\textwidth}
			\begin{tikzpicture}
				\node[NN,minimum height=1cm,minimum width=1.5cm,text width=1.5cm, align=center] (PREDICTOR) at (1,0.4) []{Predictor  \\ \scriptsize{RNN}};
				\node[NN,minimum height=1cm,minimum width=1.5cm,text width=1.5cm, align=center] (Initializer) at (1,-2) []{Initializer \scriptsize{RNN}}
				edge[post] node[right]{{$\mathbf{x}_{k_0}$}} (PREDICTOR);
				\node[sig] (U)   at (-0.4,0.4) [label=above:\small{$\mathbf{U}_{k_0+1,T}$}] {};
				\node[sig] (Y)   at (2.4 ,0.4)    [label=above:\small{$\mathbf{\tilde{Y}}_{k_0+1,T}$}] {};
				\node[sig] (u_1) at (-0.6,-2) [label=above:\small{$
							\begin{bmatrix}
								\mathbf{U}_{k_0-\tau, \tau}\\
								\mathbf{Y}_{k_0-\tau, \tau}
							\end{bmatrix}$}] {};
				\draw[->,>=latex,thick] (U) -- (PREDICTOR) ;
				\draw[->,>=latex,thick] (PREDICTOR) -- (Y) ;
				\draw[->,>=latex,thick] (u_1) -- (Initializer) ;
			\end{tikzpicture}
			\caption{RNN initializer}
			\label{fig:InitializationArch2}
		\end{subfigure}				
	\end{center}
	\caption{The two proposed initializer-predictor pairs for multi-step prediction.}
	\label{fig:InitializationArch}
\end{figure}

In Figure~\ref{fig:InitializationArch1} the block diagram of this type of the initializer-predictor pair is illustrated. The underlying assumption in this approach is that the dynamics of the RNN states, defined in~\eqref{eq:RNN_states_dynamics}, over a fixed period (i.e., the initialization length) can be approximated by a static function. The initializer network approximates that function. 

\textbf{Recurrent Initializer Network}:
Since the RNN states also possess dynamic, it is also viable to employ an RNN to model their dynamic. An RNN for initialization purpose can be a sequence-to-sequence model, $\boldsymbol{\xi}(.)$, which sequentially receives the system measurement history over the initialization length, $\tau$, and produces an output sequence, $\mathbf{X}_{k_0-\tau,\tau} $,
\begin{equation}
	\mathbf{X}_{k_0-\tau,\tau} = \boldsymbol{\xi}\big{(}\mathbf{U}_{k_0-\tau,\tau}, \mathbf{Y}_{k_0-\tau,\tau}\big{)},
	\label{eq:init_state_mapping_RNN}
\end{equation}
However, only the last element of the output sequence of the initializer network, $\mathbf{x}_{k_0}$, is used,
\begin{equation}
	\mathbf{X}_{k_0-\tau,\tau} =
		\begin{bmatrix}
			\mathbf{x}_{k_0-\tau} & \mathbf{x}_{k_0-\tau+1} &\hdots & \mathbf{x}_{k_0}
		\end{bmatrix}.
	\label{eq:init_state_mapping_RNN_output}
\end{equation}

Figure~\ref{fig:InitializationArch2} illustrates the RNN-RNN initializer-predictor pair. The initializer RNN states are set to zero. Clearly, the length of the initialization segment should be long enough to capture the dynamics of the predictor states. 

In the training process, the two networks are trained together, meaning that their weights are augmented and the gradient of the total cost in \eqref{eq:MS_Pred_Total_Cost} is calculated with respect to the augmented weight vector.

\section{Model Architectures}
\label{sec:frameworks}
Two classes of models are considered. The first class implements RNN-based black-box models. The high-level architectures of the black-box models are depicted in Figure~\ref{fig:InitializationArch}. The predictors in this figure can either be a Multi-Layer-Fully-Connected (MLFC) RNN, which essentially is an RMLP with skip connections across the network, or Long-Short-Term-Memory~\cite{Schmidhuber97} cells arranged in layers. The initializer networks, however, are either MLP or LSTM.

The second class implements a hybrid of the black-box model and a simple physics-based dynamic model of the system in a single end-to-end network. The physics-based model represents a priori knowledge about the system behaviour derived from first principles while the black-box model learns the unmodeled dynamics. The hybrid model is trained for the quadrotor vehicle in this work.

\subsection{Multi-Layer-Fully-Connected RNN}
This architecture consists of sigmoid layers which are connected in series and equipped with inter-layer (skip) connections in feedforward and feedback directions. The presence of skip connections helps to attenuate the effect of the structural vanishing gradient problem. This network is previously presented and studied for multi-step prediction of a quadrotor vehicle in~\cite{Mohajerin2014_2, Mohajerin2015_1, Mohajerin2017_1}. For an MLFC with $L$ layers, each layer $G^l$, where $l=1,...,L$, has $m^l$ inputs and $n^l$ outputs (neurons). The equations governing the dynamics of $G^l$ are
\begin{equation}
	\left\{ 
  		\begin{array}{l l}
			\mathbf{x}^l_k = \mathbf{A}^l\mathbf{y}^l_{k-1} + \mathbf{B}^l\mathbf{u}^l_k+\mathbf{b}^l\\
   			\mathbf{y}^l_k = \mathbf{f}^l\big{(}\mathbf{x}^l_k\big{)},
  \end{array} \right.
\label{eq:LRLayer}
\end{equation}
where $\mathbf{x}^l_k\in\mathbb{R}^{n^l}$ is the layer activation level, $\mathbf{y}^l_k\in\mathbb{R}^{n^l}$ is the layer output, $\mathbf{u}^l_k\in\mathbb{R}^{m^l}$ is the layer input, all at time step $k$. The matrix $\mathbf{A}^l\in\mathbb{R}^{n^l}\times\mathbb{R}^{n^l}$ is the feedback weight, $\mathbf{B}^l\in\mathbb{R}^{n^l}\times\mathbb{R}^{m^l}$ is the input weight matrix, $\mathbf{b}^l\in\mathbb{R}^{n^l}$ is a bias weight vector and $\mathbf{f}^l(.):\mathbb{R}^{n^l}\to\mathcal{R}_f^{n^l}$ is the layer activation function ($\mathbf{f}^l\in\mathcal{C}^\infty$). For an MLFC with $L$ layers the input to $G^l$ at time $k$, i.e., $\mathbf{u}^l_k$ is constructed as,
\begin{equation}
		\mathbf{u}^l_k=\Big{[} \mathbf{u}_k, \mathbf{y}^1_k, \dots, \mathbf{y}^{l-1}_k,  \mathbf{y}^{l+1}_{k-1}, \dots, \mathbf{y}^L_{k-1} \Big{]}.
	\label{eq:layerinput}
\end{equation} 

\subsection{Long-Short-Term-Memory RNN}
LSTMs, first introduced in~\cite{Schmidhuber97}, consist of~\emph{cells} which are equipped with~\emph{gates}, and are referred to as~\emph{gated} RNNs.  Gates in LSTMs are sigmoid layers which are trained to let information pass throughout the network in such a way that the gradient of the information is preserved across time. There are many versions of LSTMs~\cite{Greff2017}. The version described in~\cite{sak2014}, equipped with~\emph{peephole} connections, is used in this work. Peephole connections are connection from the cell, $\mathbf{c}(.)$, to the gates. The equations of the LSTM we use in this study are given by
\begin{equation}
	\begin{aligned}
		 \mathbf{g}^ik = & \boldsymbol{\sigma}\big{(} \mathbf{W}^i_i\mathbf{u}_k + \mathbf{W}^m_i\mathbf{m}_{k-1} + \mathbf{W}^c_i\mathbf{c}_{k-1}+\mathbf{b}_i\big{)}, \\
		 \mathbf{g}^f_k = & \boldsymbol{\sigma}\big{(} \mathbf{W}^i_f\mathbf{u}_k + \mathbf{W}^m_f\mathbf{m}_{k-1} + \mathbf{W}^c_f\mathbf{c}_{k-1}+\mathbf{b}_f\big{)}, \\
		 \mathbf{g}^o_k = & \boldsymbol{\sigma}\big{(} \mathbf{W}^i_o\mathbf{u}_k + \mathbf{W}^m_o\mathbf{m}_{k-1} + \mathbf{W}^c_o\mathbf{c}_k+\mathbf{b}_o\big{)}, \\
		 \mathbf{c}_k = & \mathbf{g}^i_k\odot \mathbf{f}\big{(}\mathbf{W}^i_c\mathbf{u}_k + \mathbf{W}^m_c\mathbf{m}_{k-1} + \mathbf{b}_c\big{)} + \mathbf{g}^f_k \odot \mathbf{c}_{k-1}, \\
		 \mathbf{m}_k = & \mathbf{g}\big{(}\mathbf{c}_k\big{)} \odot \mathbf{g}^o_k, \\
		 \mathbf{y}_k = & \mathbf{h}\big{(}\mathbf{W}_y\mathbf{m}_k+\mathbf{b}_y\big{)} = \mathbf{W}_y\mathbf{m}_k+\mathbf{b}_y.
	\end{aligned}
	\label{eq:LSTM}
\end{equation}
In this set of equations, indices $i$, $f$, $o$ and $c$ correspond to the \emph{input gate}, \emph{forget gate}, \emph{output gate} and \emph{cell}.  Gate activation functions are logistic sigmoid ($\boldsymbol{\sigma}(.)$) while the cell activation function $\mathbf{g}(.)$ and the output activation function $\mathbf{h}(.)$ are chosen by the designer.  Since the problem at hand is regression, the activation functions $\mathbf{h}(.)$ and $\mathbf{g}(.)$ are set to identity and tangent-hyperbolic function, respectively.

Note that in LSTMs, there are two types of states: cell states, $\mathbf{c}_k$, and hidden states, $\mathbf{m}_k$. In our initialization scheme, they both are treated similar to the hidden states of MLFCs. That is, they are both initialized by the initializer network.

\subsection{Hybrid model}
It is possible to incorporate prior knowledge in modeling a dynamic system with RNNs to enhance both the training convergence speed and the accuracy of the predictions. In the context of system identification, white-box models are one of the most convenient ways to represent the prior knowledge. Depending on the simplifying assumptions taken in white-box modeling, as well as the intended usage for the final model, the definition of the states and input to the white-box model may vary. Therefore, the input to the white-box model is not necessarily the measured input collected in the dataset. For instance, it is difficult to measure thrust acting on a quadrotor vehicle, and the dependency of the thrust on the motor speeds is complex in general. However, it is possible to approximate this dependency throughout the training process and employ a white-box model which receives thrust as input. Therefore, the first level of learning is to approximate the required input to the white-box model from the measured input, which is unsupervised in nature since it is already assumed that the required input to the white-box is not measured during dataset collection. 

\begin{figure}[h]
	\centering
	\resizebox{250pt}{90pt}{
	\begin{tikzpicture}
	[Box/.style={rectangle,minimum size=1.5cm,minimum height=1.3cm, thick, draw=green!50!black!50,top color=white,bottom color=green!50!black!20,inner sep=7pt}]
	  \node[NNblock,minimum height=2cm,minimum width=1.0cm] (IM) at (-5.5,0)  {\textbf{IM}};
	  \node[plantblock,minimum height=2cm,minimum width=1.0cm] (WB) at (-3.25,0)  {\textbf{PM}};
	  \node[NNblock,minimum height=2cm,minimum width=1.0cm] (OM) at (0.5,0) {\textbf{OM}};
  	  \node[Sigma]  (S1) at (1.8,0) {\large{$+$}};
 	  \node[gain, shape border rotate=-90,text width=4mm, label={center:{\footnotesize{$\mathbf{w}$}}}] (US1)  at (-1.5,0) {};
 	  \node[gain, shape border rotate=90,text width=4mm, label={center:{\small{$\mathbf{w}^{-1}$}}}] (DS2)  at (-1.15,-1.7) {};
	  
	  \node[sig]    (u)     at (-7,0)   [label=left:$\mathbf{u}$]{};
	  \node[sig]    (ths)   at (-4.35,0) [label=above left :\small{$\mathbf{r}$}]{};
	  \node[sig]    (y1)    at (-2.4,-1.7) [label=below :\small{$\tilde{\mathbf{v}}$}]{};
	  \node[sig]    (y2)    at (-2.4,0) [label=above :\small{$\hat{\mathbf{v}}$}]{};
	  \node[sig]    (y4)    at (-2.4,-0.7) [label=above :\small{${\hat{\mathbf{p}}}$}]{};
	  \node[sig]    (v_n_hat)    at (-0.55,0) [label=above :\small{$\hat{\mathbf{v}}_n$}]{};
	  \node[sig]    (v_n_tilde)    at (2.8,0) [label=above :\small{$\tilde{\mathbf{v}}_n$}]{};

	  \draw [post,line width=2pt,text width=1cm,text centered] (u) to (IM);
	  \draw [post,line width=2pt,text width=1cm,text centered] (IM) -- (ths) -- (WB);
	  
	  \draw [post,line width=2pt,text width=1cm,text centered] (WB) -- (y2) --  (US1);
	  \draw [post,line width=2pt,text width=1cm,text centered, color=blue!60] (-2.75, -0.7) -- (y4)-- (-2.4, -1.4) -- (-4.25,-1.4) -- (-4.25, -0.7) --  (-3.75, -0.7);
	  	  
	  \draw [post,line width=2pt,text width=1cm,text centered] (US1) -- (v_n_hat) -- (OM);
	  \draw [post,line width=2pt,text width=1cm,text centered] (OM) -- (S1);
	  \draw [post,line width=2pt,text width=1cm,text centered] (S1) -- (v_n_tilde) ;
	  \draw [post,line width=2pt,text width=1cm,text centered] (v_n_hat) -- (-0.55, -1.4) -- (1.8, -1.4) -- (S1) ;
  	  \draw [post,line width=2pt,text width=1cm,text centered, color=blue!60] (v_n_tilde) -- (2.8,-1.7) -- (DS2);
	  \draw [post,line width=2pt,text width=1cm,text centered, color=blue!60] (2.8,-1.7) -- (2.8,-2.4) -- (-6.7,-2.4) -- (-6.7,-0.5) -- (IM);	  
	  \draw [post,line width=2pt,text width=1cm,text centered, color=blue!60] (DS2) -- (y1) -- (-4.6,-1.7) -- (-4.6,-0.35) -- (-3.75,-0.35);
	  \draw [post,line width=2pt,text width=1cm,text centered] (u) to (-7, 0) to (-7,1.5) to (0.4, 1.5) to (OM);
	  \draw [post,line width=2pt,text width=1cm,text centered] (ths) -- (-4.35,1.2) -- (-0.9,1.2) -- (OM);
	\end{tikzpicture}}
	\caption{A suggested method to incorporate white-box models with RNNs as black box models (Hybrid or grey-box).}
	\label{fig:Hybrid}
\end{figure}
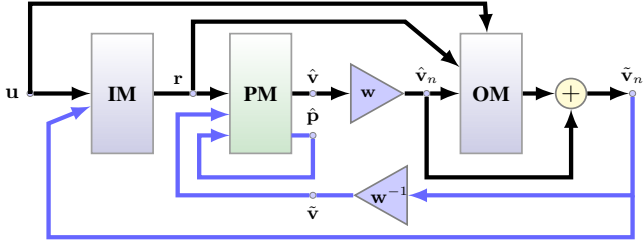

The desired values for the white-box model output, on the other hand, are partially or entirely measured during dataset collection. However, since the white-box model does not capture many of the complex nonlinearities acting on the system, its output may be too inaccurate to be useful in generating predictions of the measured output. Therefore, a second level of training is deemed necessary for compensating for the error between the white-box model prediction and the actual output measurements. Figure~\ref{fig:Hybrid} illustrates our suggested hybrid model based on the two level training discussed above. It comprises three modules; an Input Model (IM), a Physics Model (PM) and an Output Model (OM). The PM module represents the white-box model while the IM and OM modules are RNNs (with initialization networks).
The PM module receives $\mathbf{r}$, generated by the IM module, as input and updates the state vector, $\mathbf{x}=[\hat{\mathbf{p}}, \hat{\mathbf{v}}]$. The vector $\hat{\mathbf{p}}$ represents the part of the states that can be updated using integrators only, e.g., position. The vector $\hat{\mathbf{v}}$ is additionally dependent on the input to the PM module. The main responsibility of the OM module is to compensate for the deviation of the $\hat{\mathbf{v}}$ vector from the output measurements. The weights denoted by $\mathbf{w}$ normalize the input to the OM module for numerical stability. The OM module generates correction to the white-box prediction, assuming the white-box prediction error can be compensated by an additive term. The inclusion of the PM module output in the hybrid model output ensures that the PM module is actively engaged in the prediction. During the supervised training of the hybrid model, if the PM module only receives the error information through the OM module, it is possible, and highly likely, that the PM module is not effectively employed and the OM module, due to its many DOFs, learns a mapping which is too far from being a reliable and accurate model for the system~\cite{Mohajerin2017_2}.

\begin{figure}[h]
	\centering
	\resizebox{250pt}{150pt}{
	\begin{tikzpicture}

	  \node[NNblock,minimum height=3cm,minimum width=1cm] (IM) at (-6.25,0)  {\textbf{IM}};
	  \node[plantblock,minimum height=4cm,minimum width=1cm] (WB) at (-3.75,0)  {\textbf{MM}};
	  \node[NNblock,minimum height=3cm,minimum width=1cm] (OM) at (1.3,0) {\textbf{OM}};
 	  \node[gain, shape border rotate=-90,text width=5mm, label={center:{\footnotesize{$\mathbf{w}_\omega$}}}] (US1)  at (-1.5,1) {};
	  \node[gain, shape border rotate=-90,text width=5mm, label={center:{\footnotesize{$\mathbf{w}_{\dot{\xi}}$}}}] (US2)  at (-1.5,-1) {};
	  \node[Sigma]  (S1) at (2.6,0.7) {\large{$+$}};
	  \node[Sigma]  (S2) at (2.6,-0.7) {\large{$+$}};
	  
	  \node[gain, shape border rotate=90,text width=5mm, label={center:{\footnotesize{$\mathbf{w}^{-1}_\omega$}}}] (DS1)  at (-1,2.8) {};
 	  \node[gain, shape border rotate=90,text width=5mm, label={center:{\small{$\mathbf{w}^{-1}_{\dot{\xi}}$}}}] (DS2)  at (-1,-2.8) {};
	  
	  \node[sig]    (u)     at (-8,0)   [label=above:$\mathbf{u}$]{};
	  
	  \node[sig]    (ths)   at (-5.25,0) [label=below :\small{$\tilde{\boldsymbol{\tau}}$}]{};

	  \node[sig]    (y1)    at (-2.8,1.5) [label=right :\small{$\hat{\boldsymbol{\eta}}$}]{};
	  \node[sig]    (y2)    at (-2.8,0.5) [label=above :\small{$\hat{\boldsymbol{\omega}}$}]{};
	  \node[sig]    (y3)    at (-2.8,-0.5) [label=above :\small{$\dot{\hat{\boldsymbol{\xi}}}$}]{};
	  \node[sig]    (y4)    at (-2.8,-1.5) [label=right :\small{$\hat{\boldsymbol{\xi}}$}]{};
	  
	  \node[sig]    (y2_n)    at (-0.35,1) [label=below :\small{$\hat{\boldsymbol{\omega}}_n$}]{};
	  \node[sig]    (y3_n)    at (-0.35,-1) [label=above :\small{$\dot{\hat{\boldsymbol{\xi}}}_n$}]{};

	  \node[sig]    (y_1)    at (3.5,0.7) [label=below :\small{$\tilde{\boldsymbol{\omega}}_n$}]{};
	  \node[sig]    (y_2)    at (3.5,-0.7) [label=above :\small{$\dot{\tilde{\boldsymbol{\xi}}}_n$}]{};
	  
	  \draw [post,line width=2pt,text width=1cm,text centered] (u) to (IM);
	  \draw [post,line width=2pt,text width=1cm,text centered, color=blue!60] (-3.25, 1.5) -- (y1) -- (-2.8, 2.5) -- (-4.75,2.5) -- (-4.75, 1.5) -- (-4.2, 1.5);
	  \draw [post,line width=2pt,text width=1cm,text centered] (-3.25, 0.5) -- (y2) -- (-2.5, 0.5) -- (-2.5, 1) -- (US1);
	  \draw [post,line width=2pt,text width=1cm,text centered] (-3.25, -0.5) -- (y3)-- (-2.5, -0.5) -- (-2.5, -1) -- (US2);
	  \draw [post,line width=2pt,text width=1cm,text centered, color=blue!60] (-3.25, -1.5) -- (y4)-- (-2.8, -2.5) -- (-4.75,-2.5) -- (-4.75, -1.5) -- (-4.2, -1.5);
	  	  
	  \draw [post,line width=2pt,text width=1cm,text centered] (US1) -- (y2_n) --(0.8, 1);
	  \draw [post,line width=2pt,text width=1cm,text centered] (US2) -- (y3_n) --(0.8, -1);
	  
	  \draw [post,line width=2pt,text width=1cm,text centered] (y2_n) to(-0.35, 2) to (-0.15, 2) to [rounded corners] (0.15,2.2) to (0.4, 2) to (2.6, 2) -- (S1);
	  \draw [post,line width=2pt,text width=1cm,text centered] (y3_n) -- (-0.35, -2) -- (2.6, -2) -- (S2);
	  
	  \draw [post,line width=2pt,text width=1cm,text centered] (1.75, 0.7) -- (S1) ;
	  \draw [post,line width=2pt,text width=1cm,text centered] (1.75, -0.7) -- (S2);
	  
	  \draw [post,line width=2pt,text width=1cm,text centered] (S1) -- (y_1);
	  \draw [post,line width=2pt,text width=1cm,text centered] (S2) -- (y_2);
	  
	  \draw [post,line width=2pt,text width=1cm,text centered, color=blue!60] (y_1) -- (3.5,2.8) -- (DS1);
	  \draw [post,line width=2pt,text width=1cm,text centered, color=blue!60] (y_2) -- (3.5,-2.8) -- (DS2);
	  
	  \draw [post,line width=2pt,text width=1cm,text centered, color=blue!60] (y_1) to (3.5,4) to (-7.25, 4) to (-7.25, 3.85)to [rounded corners] (-7.15, 3.75) to (-7.25, 3.65) to (-7.25, 0.75) to (-6.75,0.75);
	  
	  \draw [post,line width=2pt,text width=1cm,text centered, color=blue!60] (y_2) to (3.5,-3.7) to (-7.25, -3.7) to (-7.25, -0.75) to (-6.75, -0.75);
	  
	  \draw [post,line width=2pt,text width=1cm,text centered, color=blue!60] (DS1) -- (-5, 2.8) -- (-5, 0.75) -- (-4.2, 0.75);
	  \draw [post,line width=2pt,text width=1cm,text centered, color=blue!60] (DS2) -- (-5, -2.8) -- (-5, -0.75) -- (-4.2, -0.75);
	  \draw [post,line width=2pt,text width=1cm,text centered] (IM) -- (ths) -- (WB);
	  \draw [post,line width=2pt,text width=1cm,text centered] (ths) to (-5.25,3.5) to (0, 3.5) to (0, 2.9) to [rounded corners] (0.1,2.8) to (0,2.7) to (0, 1.1)to [rounded corners] (0.1,1) to (0,0.9) to (0, -0.35) to (0.8, -0.35) ;
	  \draw [post,line width=2pt,text width=1cm,text centered] (u) to (-7.5, 0) to (-7.5,3.75) to (0.25, 3.75) to (0.25, 2.9) to [rounded corners] (0.35,2.8) to (0.25,2.7) to (0.25, 1.1)to [rounded corners] (0.35,1) to (0.25,0.9) to (0.25, 0.35) to (0.8, 0.35) ;
	\end{tikzpicture}}
	\caption{Hybrid model of a quadrotor.}
	\label{fig:Greybox_Parallel}
\end{figure}
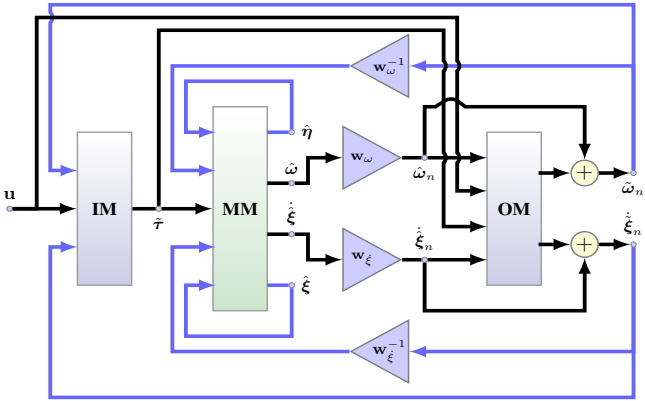
For modeling the quadrotor, the PM module implements the vehicle motion model and, therefore, is referred to as the Motion Model (MM) module. The MM module receives torques and thrust and updates the vehicle states, $\mathbf{x}$, which is,
\begin{equation}
	\label{eq:Quadrotor_States}
	\begin{aligned}
		\mathbf{x}^T &= [\boldsymbol{\eta}^T \hspace{7pt} \boldsymbol{\omega}^T \hspace{7pt} \boldsymbol{\xi}^T  \hspace{7pt} \boldsymbol{\dot{\xi}}^T]\\
					 &=[\phi \hspace{7pt} \theta \hspace{7pt} \psi \hspace{7pt} p \hspace{7pt} q \hspace{7pt} r \hspace{7pt} x \hspace{7pt} y \hspace{7pt} z \hspace{7pt} \dot{x}_I \hspace{7pt} \dot{y}_I \hspace{7pt} \dot{z}_I].
	\end{aligned}
\end{equation}
where $\boldsymbol{\eta}$ is the Euler angles, $\boldsymbol{\omega}$ is the body angular rates in the body frame (or body rates), $\boldsymbol{\xi}$ is the position and $\boldsymbol{\dot{\xi}}$ is the velocity, both in the inertial frame. The details of the model can be found in~\cite{Voos2009, Madani2008, Freddi2011}. This configuration is illustrated in Figure~\ref{fig:Greybox_Parallel}. Blue links represent feedback connections and black links represent feedforward connections. The gains $\mathbf{w}_\omega$ and $\mathbf{w}_{\dot{\xi}}$ are set to the maximum values of the vehicle body rates and velocity (Table~\ref{tab:Pelican_Maxes}).

\section{Results and Discussions}
\label{sec:results}
Two datasets are used for training RNNs in the described multi-step prediction problem. The first is the Stanford helicopter dataset\footnote{Available at \url{http://heli.stanford.edu/dataset/}} which is intended for Apprenticeship Learning~\cite{Abbeel2010}. The second dataset, which is collected for this work, encompasses various flight regimes of a quadrotor flying indoors. The quadrotor dataset is publicly available at \url{https://github.com/wavelab/pelican_dataset}.

\subsection{Quadrotor Dataset}
\label{sec:QRDataset}
The quadrotor dataset consists of time-series samples which are recovered from post-processing measurements of the states of a flying quadrotor in a cubic indoor space. The total recorded flight time is approximately 3 hours and 50 minutes, which in total corresponds to about 1.4 million time steps for each time-series. The vehicle states are measured using onboard sensors as well as a precise motion capture system. The vehicle is operated by a human pilot in various flight regimes, such as hover, slight, moderate and aggressive maneuvers. The maximum values for the 6 DOFs of the vehicle are listed in Table~\ref{tab:Pelican_Maxes}. For more details about the dataset refer to \cite{Mohajerin_PhDThesis}.
\begin{table}[htp!]
\renewcommand{\arraystretch}{0.8}
	\centering
	\setlength{\tabcolsep}{4.7pt}.
	\setlength\extrarowheight{6pt} 
	\begin{tabular}{| c | c | c | c | c | c |}		
		\hline
		 $\dot{x}$ \footnotesize{$(m/s)$} & $\dot{y}$ \footnotesize{$(m/s)$} & $\dot{z}$ \footnotesize{$(m/s)$} & $p$ \footnotesize{$(rad/s)$} & $q$ \footnotesize{$(rad/s)$}  & $r$ \footnotesize{$(rad/s)$}  \\
		\hline
		3.9268 & 3.9721 & 5.8526 & 3.9116 & 3.8506 & 3.7902\\
		\hline
	\end{tabular}
	\caption{Maximum values for the Pelican measurements.}
	\label{tab:Pelican_Maxes}
\end{table}

\subsection{Helicopter Dataset}
The Helicopter data set was collected in August 2008 as a part of research for Apprenticeship Learning~\cite{Abbeel2010}. It has also been used for a single-step prediction system identification problem~\cite{Abbeel2015}. The flights are carried out in an outdoor environment, however, the dataset does not provide a wind measurement. The flight time is approximately 55 minutes and there are 300k samples for each quantity. For more information about the dataset see \url{http://heli.stanford.edu/index.html}.

\subsection{Architectures and Implementation}

The predictor network can be either an MLFC, LSTM or LSTM equipped with buffers. The buffers are called Tapped Delay Lines (TDLs)~\cite{Haykin_Book_1999}. The TDL size is 10, throughout the experiments in this work. Each of the predictors may be initialized in one of the three fashions: washout, with an MLP initializer or with an RNN initializer. In case of an RNN initializer, an LSTM with one layer of LSTM cells is employed. In order to refer to each configuration, the following notation is used:
\begin{center}
	[predictor]: [number of layers] $\times$ [size of each layer] - [initializer type]: [hidden layer size]$\times$[initialization length]
\end{center}
For example, an LSTM predictor with 3 layers, each having 200 LSTM cells initialized by an MLP with 1000 neurons in the hidden layer and an initialization length of 10 is referred to by \small \textbf{LSTM: 3$\times$200-MLP:1000$\times$10} \normalsize.

The initialization length is 0.1s, or 10 steps, for all of the networks trained in this work. Two prediction lengths are considered, $T_{pred}=0.4$s and $T_{pred}=1.9$s, which correspond to 40 steps and 190 steps of prediction, respectively.

\subsection{Evaluation Metrics}
To evaluate the network prediction accuracy, the mean and distribution of the prediction error at each time step are studied. The distributions will be illustrated using box-plots over the two aforementioned prediction lengths. Note that for the evaluation the test dataset is used, i.e., each sample for evaluation has not been used to train the network. The following norms are used,
\begin{equation}
    \begin{aligned}
    ||\bar{e}_{\dot{\xi},k}|| =& \tfrac{1}{3}\big{(}|e_{\dot{x}_I,k}|+|e_{\dot{y}_I,k}|+|e_{\dot{z}_I,k}|\big{)},\\
    ||\bar{e}_{\omega,k}|| =& \tfrac{1}{3}\big{(}|e_{p,k}|+|e_{q,k}|+|e_{r,k}|\big{)},\\
    ||\bar{e}_{\dot{\eta},k}|| =& \tfrac{1}{3}\big{(}|e_{\dot{\phi},k}|+|e_{\dot{\theta},k}|+|e_{\dot{\psi},k}|\big{)},
    \end{aligned}
\end{equation}
where $\bar{e}_{\dot{\xi},k}$ is measured in meters per second (m/s) and $\bar{e}_{\omega,k}$ and $\bar{e}_{\dot{\eta},k}$ are measured in degrees per second (deg/sec). The subscripts, $\dot{\xi}$, $\omega$ and $\dot{\eta}$ correspond to the velocity, body rates and Euler rates, respectively. Note that each error is averaged over the three perpendicular axes to give the average errors on each component of each vector. All the three prediction errors are calculated at each prediction step, $k$. Note that using absolute values weighs all errors equally, regardless of sign. Using other norms does not significantly alter the comparative results. 

\vspace{-5pt}
\subsection{NN-based Initialization vs. Washout}
\begin{figure}[h!]
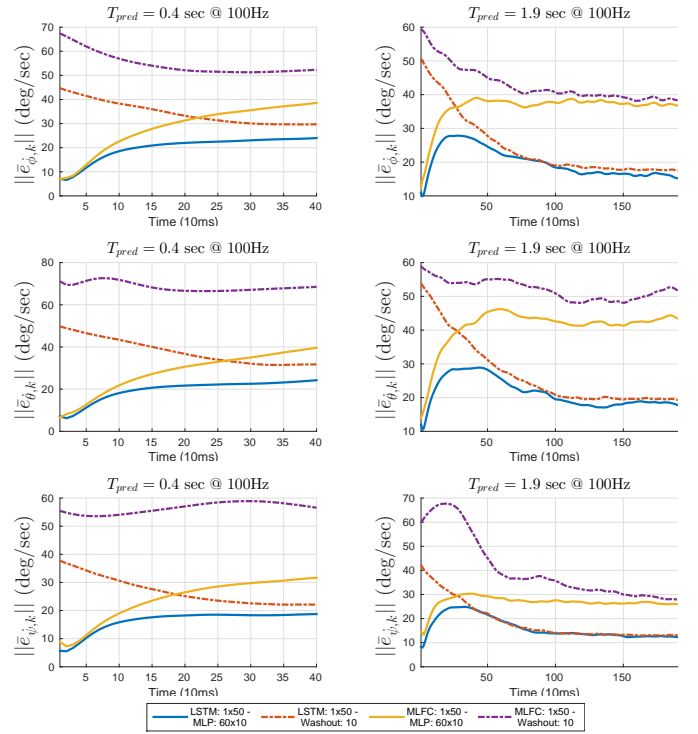

        \centering
		\includegraphics[scale=0.42,trim= 0mm 1mm 0mm 0mm,clip]{figures/results/helicopter/inertial_roll_mlp_vs_washout.eps}
		\includegraphics[scale=0.42,trim= 0mm 1mm 0mm -5mm,clip]{figures/results/helicopter/inertial_pitch_mlp_vs_washout.eps}
		\includegraphics[scale=0.42,trim= 0mm 1mm 0mm -5mm,clip]{figures/results/helicopter/inertial_yaw_mlp_vs_washout.eps}
		\includegraphics[scale=0.42,trim= 60mm 0mm 0mm 47mm,clip]{figures/results/Legend.eps}
	\caption{Comparison between NN-based (using MLP) and washout initialization on the simplified helicopter dataset. From top to bottom: roll rate, pitch rate and yaw rate.}
	\label{fig:HeliCompare}
\end{figure}

To compare the effect of our NN-based initialization method versus washout, simple configurations of MLFCs and LSTMs, initialized by either our method or washout, are trained and studied. To save training time, the networks are trained on three subsets of the helicopter dataset. Each dataset belongs to a Multi-Input-Single-Output (MISO) subsystem of the helicopter which maps the pilot commands to the Euler rates. The following architectures are trained and compared in Figure~\ref{fig:HeliCompare}:

\small
\begin{itemize}
	\item \textbf{MLFC: 1$\times$50 - MLP: 60$\times$10}
	\item \textbf{MLFC: 1$\times$50 - Washout: 10}
	\item \textbf{LSTM: 1$\times$50 - MLP: 60$\times$10}
	\item \textbf{LSTM: 1$\times$50 - Washout: 10}
\end{itemize}
\normalsize

From Figure~\ref{fig:HeliCompare} it is observed that the NN-based initialization method has improved both the immediate and the overall prediction accuracy. When the prediction error is long, however, the NN-based and washout initialized RNN predictors converge to almost the same error eventually. It is also evident that an efficient washout period is difficult to determine, whereas in our NN-based initialization methods such a problem does not exist. The results clearly show the superiority of our NN-based initialization scheme over the state-of-the-art.

\vspace{-15pt}
\subsection{MLFC vs. LSTM}
Having established the superiority of NN-based initialization scheme, next the predictor architecture is evaluated. The following architectures are trained on the same subsets of the helicopter dataset: 

\small
\begin{itemize}
	\item \textbf{MLFC: 1$\times$50 - MLP: 60$\times$10}
	\item \textbf{MLFC: 2$\times$50 - MLP: 100$\times$10}
	\item \textbf{MLFC: 2$\times$100 - MLP: 200$\times$10}
	\item \textbf{LSTM: 1$\times$50 - MLP: 60$\times$10}
	\item \textbf{LSTM: 2$\times$50 - MLP: 100$\times$10}
\end{itemize}
\normalsize

As a comparative measure, the following Root Mean Sum of Squared Errors (RMSSE) measure is calculated on the test dataset,
\begin{equation}
	RMSSE = \sqrt{\frac{1}{T_{pred}n_\mathcal{G}}\sum_{i=1}^{n_\mathcal{G}}\sum_{k=\tau+1}^{T_{pred}}\mathbf{e}_i^\top(k)\mathbf{e}_i(k)},
\end{equation}
where $n_\mathcal{G}$ is the size of the test dataset, $\mathcal{G}$.

\begin{figure}[htp!]
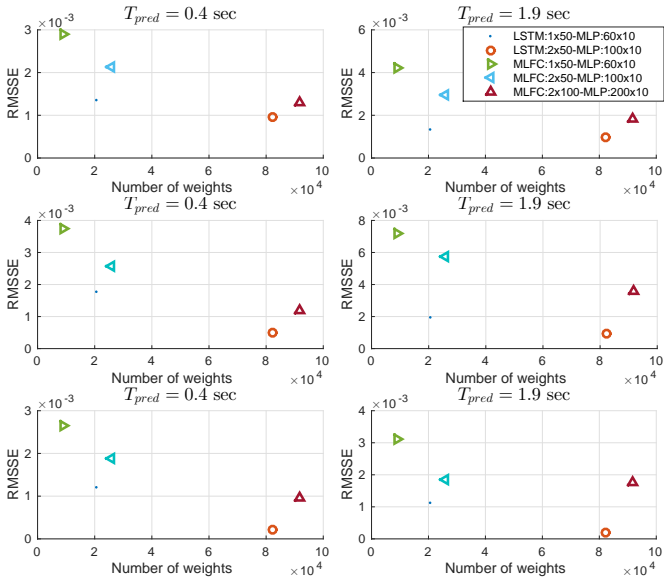

	\begin{center}
		\includegraphics[scale=0.485,trim= 0mm 0mm 0mm 0mm,clip]{figures/results/helicopter/size_compare_roll_rate.eps}	
		\includegraphics[scale=0.485,trim= 0mm 0mm 0mm 0mm,clip]{figures/results/helicopter/size_compare_pitch_rate.eps}
		\includegraphics[scale=0.485,trim= 0mm 0mm 0mm 0mm,clip]{figures/results/helicopter/size_compare_yaw_rate.eps}
	\end{center}
	\vspace{-10pt}
	\caption{Network size vs. $RMSSE$ and LSTMs vs. MLFCs, helicopter reduced dataset.}
	\label{fig:HeliSizeCompare}
\end{figure}

In Figure~\ref{fig:HeliSizeCompare} the RMSSE measure versus the size of the networks (number of weights) are plotted for the aforementioned architectures. In these graphs it is observed that the LSTMs with MLP initialization outperform other methods. In fact, LSTMs with fewer weights perform better than MLFCs. As a part of hyper-parameter optimization, it is a reasonable choice to conduct the remaining experiments with LSTMs initialized by the NN-based scheme.

\vspace{-10pt}
\subsection{MLP vs RNN Initializers}
In this section, variants of LSTM networks initialized with the two initializer networks are examined. The LSTM networks are comprised of layers of LSTM cells connected in series. The last layer output is fed back to the first layer. For some experiments, to provide the predictor with a truncated history of the signals, TDLs are placed at the input and output of the networks. The networks map the pilot commands to the Euler rates (4 inputs and 3 outputs). The following architectures are trained and assessed:

\small
\begin{itemize}
	\item \textbf{LSTM: 7$\times$200 - MLP: 15000$\times$10}
	\item \textbf{LSTM: 7$\times$200 - RNN: 2500$\times$10}
	\item \textbf{LSTM TDL: 7$\times$200 - MLP: 15000$\times$10}
	\item \textbf{LSTM TDL: 7$\times$200 - RNN: 2500$\times$10}
\end{itemize}
\normalsize

\vspace{-5pt}
\begin{figure}[htp!]
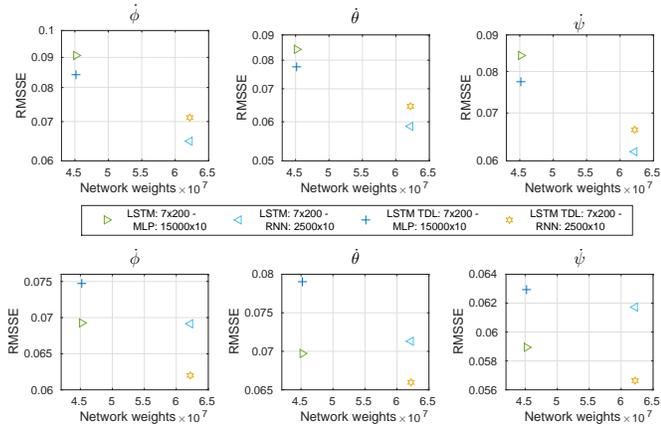

	\begin{center}
		\hspace{3pt}\includegraphics[scale=0.44,trim= 0mm 0mm 0mm 0mm,clip]{figures/results/helicopter/angular_rates_full_size_compare_50.eps}\\
		\vspace{5pt}
		\includegraphics[scale=0.44,trim= 0mm 0mm 0mm 0mm,clip]{figures/results/helicopter/angular_rates_full_size_compare_200.eps}
	\end{center}
	
	\caption{Comparisons of network sizes and types on learning the helicopter angular rates from pilot commands. (Top row: $T_{pred}=0.4$s, bottom row: $T_{pred}=1.9$s)}
	\label{fig:HeliSizeCompareEuler}
\end{figure}

Figure~\ref{fig:HeliSizeCompareEuler} illustrates the total RMSSE cost on the test dataset for the helicopter angular rates over the previously mentioned prediction lengths. The networks are quite large, therefore, a few strategies were chosen to prevent overfitting; the network weights were initialized to tiny numbers, weight decay regularizer and drop-out method~\cite{Srivastava2014} were also employed. Since the measures are calculated over the test dataset, overfitting did not occur. Based on the results illustrated in Figure~\ref{fig:HeliSizeCompareEuler}, it can be seen that the trained models behave similarly and therefore, we mainly focus on these four architectures.

\subsection{Black-box Modeling of the Helicopter}
To study the reliability and accuracy of the predictions it is best to look at the~\emph{distribution} of the prediction error, across the test datasets, throughout the prediction length.

\begin{figure}[ht!]
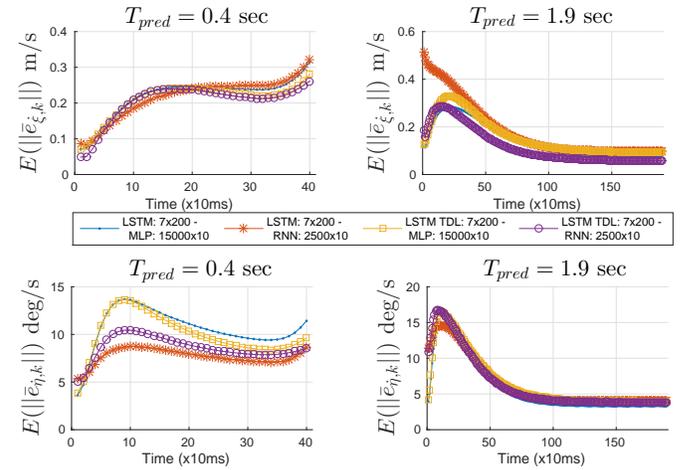

	\begin{center}
		\includegraphics[scale=0.48,trim= 0mm 0mm 0mm 0mm,clip]{figures/results/helicopter/distributions/vel/mean.eps} \\
		\vspace{5pt}
		\includegraphics[scale=0.48,trim= 0mm 0mm 0mm 0mm,clip]{figures/results/helicopter/distributions/ang_rate/mean.eps}
	\end{center}

	\caption{Mean of the error distributions for the four black-box models of the helicopter. The top row plots correspond to the velocity and the bottom row plots correspond to angular rates.}
	\label{fig:HeliMeanCompare}
\end{figure}

Figure~\ref{fig:HeliMeanCompare} compares the mean of the error distributions over the two prediction lengths. It would have been expected that the prediction error increases monotonically throughout the prediction length. This is more or less the case for $T_{pred} = 0.4$s. However, for longer prediction lengths the monotonic increasing behaviour is no longer observed. Instead, a peak appears at the early stages of the prediction and it is attenuated as we go forward in time. This is contrary to our expectation. 

As the LSTMs are efficient in learning long-term dependencies~\cite{Hochreiter2001}, the decrease in the error over late predictions might be due to noise attenuation and accumulating more information about the process by the LSTMs. Also, the peaks may be reduced if the initialization length increases. However, increasing the initialization length decreases the lengths to be used for training the predictor networks. It is also observed that the LSTMs, equipped by TDLs and initialized by RNNs generally perform better for longer horizons, which reinforces this hypothesis. However, for the Euler rate predictions, the behaviour of the mean error is not consistent. This inconsistency may be due to the size and quality of the helicopter dataset. Some of the drawbacks in using the helicopter dataset for multi-step prediction are,

\begin{enumerate}
	\item The input to the networks is the pilot command and there are many levels of transformation which take place before the commands affect the helicopter motion. Time synchronization can also become very difficult to manage in such situations. To circumvent this, using actual motor speeds as the inputs is likely to mitigate effects of delay and command transformation.
	\item Considering the complex dynamics of a helicopter, the dataset is relatively small. Better prediction performance is expected if more data is collected in a variety of flight regimes.
	\item The helicopter is flown outdoors and is very likely affected by wind. However, no information about the wind is provided in the helicopter dataset. To obtain a predictor for the vehicle dynamic a controlled environment is more desirable.
\end{enumerate}

As we have considered the above drawbacks in collecting the quadrotor dataset, it is expected that quadrotor dataset better suits the multi-step prediction problem at hand.

\subsection{Black-box Modeling of the Quadrotor}
\label{seq:BB_Quadrotor}
The quadrotor models in this study map a trajectory of the motor speeds and a truncated history of the system states (to initialize the RNNs) to the vehicle velocity and body angular rate vectors (or body rates). Learning velocity directly from motor speeds is difficult, because of the dependency of the velocity on the vehicle attitude. A network whose output comprises of both the velocity and body rates is difficult to train, as the network output associated with the velocity diverges during early stages of the training and prohibits the training to converge. Therefore, as proposed in~\cite{Mohajerin2015_1}, the velocity and body rates are decoupled and learned separately: one network learns to predict body rates from motor speeds and the other learnes to predict the velocity from motor speeds and \emph{body rates}. As the training is offline, the actual body rates are used to train the second network, which resembles the \emph{teacher forcing} method~\cite{Narendra90}. However, to use the two networks for multi-step prediction, the first network provides the second network with the \emph{predicted} body rates, in a cascaded architecture. We call this mode \emph{practical}. 

\begin{figure}[htp!]
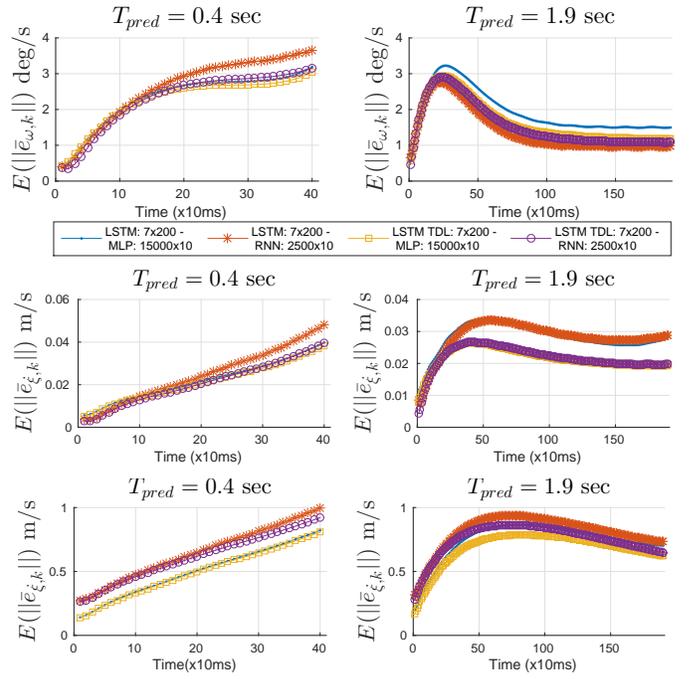

	\begin{center}
		\includegraphics[scale=0.50,trim= 0mm 0mm 0mm 0mm,clip]{figures/results/quadrotor/distributions/pqr/BB/mean.eps}\\
		\vspace{5pt}
		\includegraphics[scale=0.48,trim= 0mm 0mm 0mm 0mm,clip]{figures/results/quadrotor/distributions/vel/BB/mean.eps} \\
		\vspace{5pt}
		\includegraphics[scale=0.48,trim= 0mm 0mm 0mm 0mm,clip]{figures/results/quadrotor/distributions/vel/BB/L1_practical/mean.eps} 
	\end{center}
	\caption{Mean of the error norms for the black-box model of the quadrotor, over the two prediction lengths. From top to bottom, the y axis corresponds to body rates, velocity in the teacher force and velocity in the practical mode.}
	\label{fig:QuadMeanCompare}
\end{figure}

Figure \ref{fig:QuadMeanCompare} illustrates the mean of the error norms, measured at each prediction step, for the aforementioned architectures, over the two prediction lengths. For the body rate prediction (the top row), the prediction accuracy on average remains better than 3.5 (deg/sec) over 1.9 seconds prediction length ($T_{pred}=1.9s$). For this length, similar to the helicopter black-box model, an initial increase in the prediction error is observed. Although the error improves as the prediction proceeds, this initial increase is not favourable in a control application. The later improvement of the prediction error is due to the properties of the LSTM network as discussed for the helicopter model.

The plots in the middle row illustrate the average of the norm of the velocity prediction errors in the teacher force mode, i.e., the samples in the test dataset include the measured body rates as inputs. The accuracy of the predictions on average remains better than 4 cm/s over almost 2 seconds. From the plots on the first and second rows, it is also observed that TDLs improve the prediction accuracy, which is consistent with our observation from the helicopter dataset. It is also observed that the LSTMs initialized with RNNs (RNN-RNN pairs) have better prediction accuracy over the longer prediction lengths; a reinforcing observation on the argument that the LSTMs efficiently employ information spread across time. 

The bottom row corresponds to the velocity prediction errors in the practical mode. Comparing the teacher force mode, the velocity prediction accuracy is degraded by a factor of approximately 25. Additionally, it can be observed that the networks with the RNN initializer suffer more from the error in body rate prediction. In conclusion, the black-box model provides a reliable and accurate body rate prediction, however, the velocity prediction is far from desirable. 

%

\subsection{Hybrid Mmodel of the Quadrotor}
As opposed to the black-box model, the hybrid model (Figure~\ref{fig:Greybox_Parallel}) provides the velocity and body rate prediction simultaneously. The IM and OM modules in the hybrid model are initialized RNNs. The proceeding results correspond to the IM and OM modules configured as \small \textbf{LSTM: 4$\times$200 - MLP: 5000$\times$10}\normalsize. In this section, the error distributions are studied to further assess the model prediction accuracy.

\begin{figure}[h]
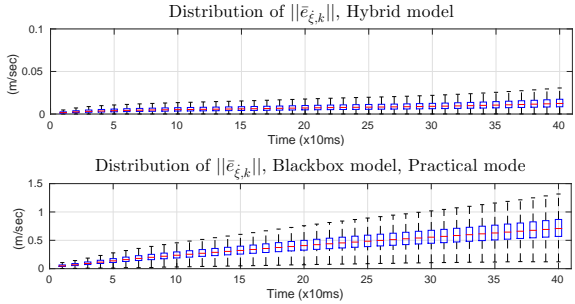

	\begin{center}
		\includegraphics[scale=0.420,trim= 0mm 0mm 0mm 0mm,clip]{figures/results/quadrotor/distributions/vel/Hybrid/T50.eps}\\
		\vspace{5pt}
		\hspace{2pt}\includegraphics[scale=0.420,trim= 0mm 0mm 0mm 0mm,clip]{figures/results/quadrotor/distributions/vel/BB/L1_practical/T50.eps}
	\end{center}
	\vspace{-5pt}
	\caption{Velocity error distribution of the hybrid (top) and the black-box (bottom) models ($T_{pred}=0.4$s). Note that the prediction by the hybrid model is over an order of magnitude better than the black-box.}
	\label{fig:GB_BB_Dist_Vel50}
\end{figure}

In Figures~\ref{fig:GB_BB_Dist_Vel50},~\ref{fig:GB_BB_Dist_pqr50},~\ref{fig:GB_BB_Dist_Vel200} and~\ref{fig:GB_BB_Dist_pqr200} we compare the error distributions of the black-box model (in the practical mode) and the hybrid model for the two prediction lengths. For the shortest prediction length ($0.4s$) the hybrid model reduces the body rate error by $50\%$ and the velocity error by about $20\times$ compared to the black-box model.

For the body rate prediction over $T_{pred}=1.9$s, the black-box initially performs worse, however, in the long run it performs better than the hybrid model. One possible explanation for this behaviour is that the MM module in the hybrid model limits the exploration capacity of the OM module. Although, the inclusion of the motion model contributes significantly to the accuracy of the early prediction steps, it may also impose restrictions on the search for the optimal weights. However, the LSTMs in the black-box model are free to explore the entire weight space and they can accumulate relevant information over longer time to perform better. For the velocity, we see that black-box still performs worse, however, a slight decrease is seen on the later predictions which is similar to the behaviour in the body rate prediction.

\begin{figure}[h]
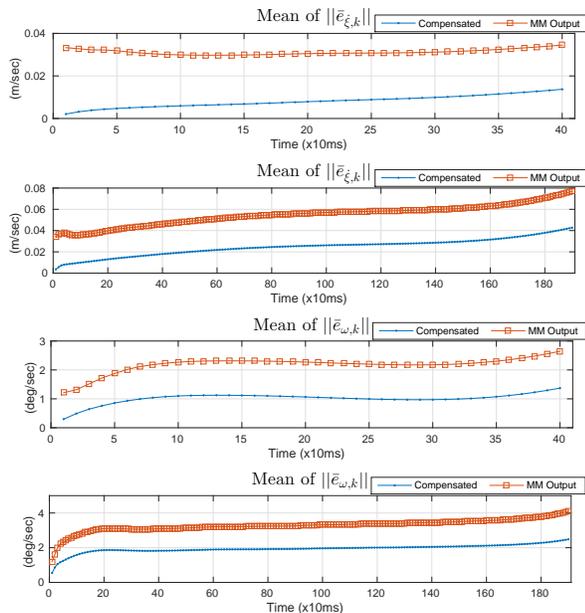

	\begin{center}
		\includegraphics[scale=0.420,trim= 0mm 0mm 0mm 0mm,clip]{figures/results/quadrotor/WB_Compensated_Vel_T50.eps}\\
		\vspace{5pt}
		\includegraphics[scale=0.420,trim= 0mm 0mm 0mm 0mm,clip]{figures/results/quadrotor/WB_Compensated_Vel_T200.eps}\\
		\vspace{5pt}
		\hspace{5pt}\includegraphics[scale=0.420,trim= 0mm 0mm 0mm 0mm,clip]{figures/results/quadrotor/WB_Compensated_pqr_T50.eps}\\
		\vspace{5pt}
		\hspace{4pt}\includegraphics[scale=0.420,trim= 0mm 0mm 0mm 0mm,clip]{figures/results/quadrotor/WB_Compensated_pqr_T200.eps}\\
	\end{center}
	\vspace{-10pt}
	\caption{The means of uncompensated output of the MM module vs. the compensated output for the velocity and body rate predictions.}
	\label{fig:MM_Dis}
\end{figure}

In Figure~\ref{fig:MM_Dis}, the mean of the prediction errors are illustrated before compensation (the MM output) and after, for each prediction lengths. This figure shows the importance of RNN network initialization. The plots show that the MM module prediction error starts from a non-zero value which requires the OM module to immediately compensate for that. Therefore, the output of the OM module should start from a non-zero value which requires a proper state initialization for the RNNs. Note that at each prediction step, the compensated states are fed back into the MM module.

\begin{figure}[h]
	\begin{center}
		\includegraphics[scale=0.420,trim= 0mm 0mm 0mm 0mm,clip]{figures/results/quadrotor/distributions/pqr/Hybrid/T50.eps}\\
		\vspace{5pt}
		\includegraphics[scale=0.420,trim= 0mm 0mm 0mm 0mm,clip]{figures/results/quadrotor/distributions/pqr/BB/T50.eps}
	\end{center}
	\caption{Comparison of the body rate error distribution between the hybrid (top) and the black-box (bottom) models ($T_{pred}=0.4$s).}
	\label{fig:GB_BB_Dist_pqr50}
\end{figure}

\begin{figure}[h]
	\begin{center}
		\includegraphics[scale=0.420,trim= 0mm 0mm 0mm 0mm,clip]{figures/results/quadrotor/distributions/vel/Hybrid/T200.eps}\\
		\vspace{5pt}
		\hspace{2pt}\includegraphics[scale=0.420,trim= 0mm 0mm 0mm 0mm,clip]{figures/results/quadrotor/distributions/vel/BB/L1_practical/T200.eps}
\end{center}
	\caption{Comparison of the velocity error distribution between the hybrid (top) and the black-box (bottom) models ($T_{pred}=1.9$s).}
	\label{fig:GB_BB_Dist_Vel200}
	\vspace{-10pt}
	
\end{figure}

\begin{figure}[h]
	\begin{center}
		\includegraphics[scale=0.420,trim= 0mm 0mm 0mm 0mm,clip]{figures/results/quadrotor/distributions/pqr/Hybrid/T200.eps}\\
		\vspace{5pt}
		\includegraphics[scale=0.420,trim= 0mm 0mm 0mm 0mm,clip]{figures/results/quadrotor/distributions/pqr/BB/T200.eps}
	\end{center}
	\vspace{-10pt}
	\caption{Comparison of the body rate error distribution between the hybrid (top) and the black-box (bottom) models ($T_{pred}=1.9$s).}
	\label{fig:GB_BB_Dist_pqr200}
\end{figure}

\vspace{-15pt}

\section{Conclusion}
\label{sec:conclusion}
In this paper, we employ RNNs to identify the dynamics of two aerial vehicles. The importance of RNN state initialization, as well as the effectiveness of our proposed method to properly initialize the states of RNNs are demonstrated. The RNNs with our proposed state initialization method are used as black-box models to learn the model of a helicopter and a quadrotor, for multi-step prediction, from experimental dataset. To overcome drawbacks from a pure black-box model, a simplified model of the quadrotor motion is embedded with RNNs in a hybrid model. The accuracy of the trained hybrid model for multi-step prediction is demonstrated experimentally. The prediction provided by the hybrid model can be effectively employed in a variety of control schemes. We hope that the provided study and experiments in this work illustrate a number of efficient ways to benefit from the ever rising power of machine learning methods in modeling and control of robotic systems.
\ifCLASSOPTIONcaptionsoff
  \newpage
\fi



%
\bibliographystyle{IEEEtran}
\bibliography{mybibfile}

\end{document}